\documentclass[journal]{IEEEtran}

\usepackage{hyperref}
\usepackage{graphicx}
\usepackage{subfigure}
\usepackage{amsmath}
\usepackage{booktabs}
\usepackage{comment}
\usepackage{amssymb}
\usepackage{bm}
\usepackage{url}
\usepackage{multirow}
\usepackage{array}
\usepackage{threeparttable}
\usepackage{physics}
\usepackage[ruled]{algorithm2e} 
\usepackage[noend]{algpseudocode}
\usepackage{color}
\usepackage{makecell}
\usepackage{array}
\usepackage{colortbl}
\usepackage{rotating}
\usepackage[normalem]{ulem}
\DeclareMathOperator*{\argmin}{argmin}
\usepackage{nomencl}
\makenomenclature

\usepackage{fancyhdr}

\begin{document}

\title{Automatic Assessment of Full Left Ventricular Coverage in Cardiac Cine Magnetic Resonance Imaging with Fisher-Discriminative 3D CNN}

\author{Le Zhang*,
        Ali Gooya,
        Marco Perea\~nez,
        Bo Dong, 
        Stefan K. Piechnik, \\
        Stefan Neubauer,
        Steffen E. Petersen,
        Alejandro F. Frangi, \textit{Fellow, IEEE}
        
        \thanks{Le Zhang (corresponding author: le.zhang@sheffield.ac.uk), Ali Gooya and Bo Dong are with the Centre for Computational Imaging and Simulation Technologies in Biomedicine (CISTIB), University of Sheffield, UK.
        	
        Marco Perea\~nez and Alejandro F. Frangi are with the Centre for Computational Imaging and Simulation Technologies in Biomedicine (CISTIB), School of Computing and School of Medicine, University of Leeds, UK.
        
        Stefan K. Piechnik and Stefan Neubauer are with the Oxford Centre for Clinical Magnetic Resonance Research (OCMR), Division of Cardiovascular Medicine, University of Oxford, John Radcliffe Hospital, UK. 
        
        Steffen E. Petersen is with Cardiovascular Medicine at the William Harvey Research Institute, Queen Mary University of London and Barts Heart Centre, Barts Health NHS Trust, UK.}
}

\maketitle

\thispagestyle{fancy}
\fancyhead{}
\lhead{}
\cfoot{\footnotesize{Copyright (c) 2017 IEEE. Personal use of this material is permitted. However, permission to use this material for any other purposes must be obtained from the IEEE by sending an email to pubs-permissions@ieee.org.}}

\begin{abstract}
Cardiac magnetic resonance (CMR) images play a growing role in the diagnostic imaging of cardiovascular diseases. Full coverage of the left ventricle (LV), from base to apex, is a basic criterion for CMR image quality and necessary for accurate measurement of cardiac volume and functional assessment. Incomplete coverage of the LV is identified through visual inspection, which is time-consuming and usually done retrospectively in the assessment of large imaging cohorts. This paper proposes a novel automatic method for determining LV coverage from CMR images by using Fisher-discriminative three-dimensional (FD3D) convolutional neural networks (CNNs). In contrast to our previous method employing 2D CNNs, this approach utilizes spatial contextual information in CMR volumes, extracts more representative high-level features and enhances the discriminative capacity of the baseline 2D CNN learning framework, thus achieving superior detection accuracy. A two-stage framework is proposed to identify missing basal and apical slices in measurements of CMR volume. First, the FD3D CNN extracts high-level features from the CMR stacks. These image representations are then used to detect the missing basal and apical slices. Compared to the traditional 3D CNN strategy, the proposed FD3D CNN minimizes within-class scatter and maximizes between-class scatter. We performed extensive experiments to validate the proposed method on more than 5,000 independent volumetric CMR scans from the UK Biobank study, achieving low error rates for missing basal/apical slice detection (4.9\%/4.6\%). The proposed method can also be adopted for assessing LV coverage for other types of CMR image data.

\end{abstract}


\IEEEpeerreviewmaketitle

\section{Introduction}
Left ventricular (LV) cardiac anatomy and function are widely used in the field of cardiac medicine for diagnosis and monitoring disease progression and for assessing the patient's response to cardiac surgery and interventional procedures. Cardiac ultrasound (US) and cardiac magnetic resonance (CMR) imaging are arguably the most widespread techniques for diagnostic imaging of the heart. For population imaging studies, however, CMR remains the modality of choice. CMR is a single technique that provides access to cardiac anatomy and non-invasive measurements of cardiac function \cite{petersen:2013}. In large population imaging studies or assessment of patient cohorts from large clinical trials, the quantification of LV anatomy and function requires automatic image quality assessment and tools for image analysis. One basic criterion for cardiac image quality is LV coverage and detection of missing apical and basal CMR slices \cite{Klinke:2013}. CMR may display incomplete LV coverage because of insufficient radiographer experience in planning a scan, natural cardiac muscle contraction, breathing motion, and imperfect triggering, all of which pose challenges in efforts at quantitative LV characterisation and accurate diagnosis \cite{Pusey:1986}. For example, missing basal slices affect calculations of LV volume and derived LV functional measures such as ejection fraction and cardiac output. Even if scout images are acquired, in order to centre the LV in view and minimize this issue, incomplete coverage may result at any point throughout the cardiac cycle because of changes in patient breathing and cardiac motion. 
Image quality assessment is traditionally performed by radiographers who ensure that patients do not leave the scanner without providing diagnostically interpretable data. However, there are limits to human attention. With CMR examinations becoming less expensive and increasingly commissioned, scanning loads at some centres may be insufficient to maintain consistent standards. Quality assessment is of particular importance in large-scale population imaging studies, where data are acquired across different imaging sites before core lab analysis. For example, large volumes of data may be stored without being checked by experienced staff prior to analysis \cite{Ferreira:2013} \cite{Wang:2006}. Automatic methods for these repetitive quality assurance tasks provide the required consistency and reliability. 

To ensure consistent quantification of CMR data, automatic assessment of complete LV coverage is the first step. LV coverage is assessed by visual inspection of CMR image sequences, which is a subjective, repetitive, error-prone, and time-consuming process \cite{Attili:2010}. Automatic coverage assessment is required to promptly intervene and correct data acquisition, and/or discard images with incomplete LV coverage whose analysis would otherwise impair any statistics aggregated over the cohort. The most common causes of incomplete LV coverage are lack of a basal slice (no atrial chamber visible in end-systole, hence no certainty that the base of the heart is completely covered) and lack of an apical slice (LV cavity remains visible at end-systole). According to the criteria used in \cite{Klinke:2013} for CMR quality assessment, a missing basal slice carries a higher penalty than a missing apical slice, given its impact on LV volume computation. Although technological developments in magnetic resonance imaging (MRI) hardware and pulse sequences have led to faster CMR acquisitions, challenges remain with regard to ensuring full heart coverage and motion compensation. In the UK Biobank's CMR protocol, for instance, incomplete heart coverage is the reason for flagging $4\%$ of all CMR examinations as providing unreliable or non-analysable image data \cite{carapella2016towards}. While 4\% may seem to be a small proportion, the challenge is to automatically sift through the entire database to identify and exclude those cases from further quantitative analysis. Methods for the objective detection of basal and apical imaging planes are relevant in this context, as their absence affects diagnostic accuracy as well as anatomical and functional LV quantification.

In the field of video processing, Automatic Image Quality Assessment (AIQA) is a well-developed corpus of techniques concerned with detecting image distortions characteristic of multimedia communications \cite{Saad:2012} \cite{Xue:2014}. These distortions generally differ from those affecting medical images. No-reference-based image quality assessment (NR-IQA) \cite{He:2012} \cite{Moorthy:2011} is relevant for medical imaging data. While there is relatively easy access to abundant data sets of mixed quality, it is not possible to collect data without some level of image degradation or artefacts. Practical CMR image-processing applications do not provide perfect versions of incomplete LV coverage images, but rather, only the image to be assessed. While assessments attempt to highlight differences in our assessed data set regarding a hypothetical high-quality image \cite{Kang:2014}, the final image quality is estimated solely based on the characteristics of the assessed image.  

The current standard operating procedure in the UK Biobank, for instance, involves the detection of missing basal/apical slices based on visual assessment by experts. Few methods have been developed for automating this process, and prior work mostly adopted approaches that require segmenting short-axis slices of LV \cite{bernard2016standardized} \cite{zhen2016multi} or landmark localization \cite{hoffmann2016automated} \cite{lu2016robust} \cite{de2017convnet}. However, fast full LV coverage detection as the first step of an image quantification pipeline is largely unexplored. Hoffmann et al. pioneered this field \cite{hoffmann2016automated} by initially localizing the heart in raw data prior to applying computer-aided diagnosis algorithms. Lu et al. \cite{lu2011automatic} proposed an approach to locate LV and prescribe long/short-axis views before MR image acquisition, which could be used to evaluate cardiac coverage in short-axis views. These methods detect missing basal/apical slices and largely rely on the quality of LV segmentation and localization. de Vos et al. \cite{de2016automatic} proposed a method that automatically identifies a slice of interest (SOI) in 3D images. A ConvNet regressor was trained to determine the distance between each 2D slice and the SOI. However, this solution does not consider 3D contextual information contained across slices. 

\begin{figure}	
	\centering	
	\includegraphics[scale=0.74]{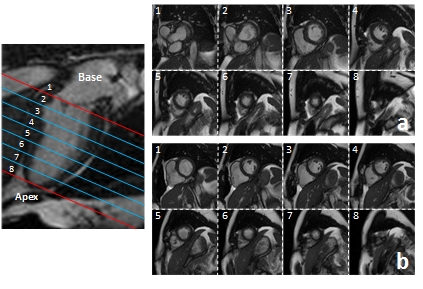} 
	\caption{\textit{Left:} A typical two-chamber view cardiac MRI with eight slices covering from base to apex. \textit{Right:} (a) a volume with whole coverage (slice 1 is the basal slice), and (b) a volume with missing basal slice (slice 1 is not the basal slice). In each rectangle, from top to bottom, rows correspond to adjacent axial slices.}\vspace{-1em}
	\label{figure 1}
\end{figure}

The characteristics of the LV are useful in identifying the position that the slice belongs to, since the LV in each slice shows a different shape and size. For example, the LV shape is approximately circular in mid-slices, while it is more elliptical in basal slices (Fig. \ref{figure 1}). Recent work \cite{Fouhey:2016}, \cite{Xie:2016} has focused on learning data-driven features to more accurately detect shape differences. Among them, 3D convolutional neural networks (CNNs) are one of the most regularly used deep-learning schemes to meet the challenges of discriminative shape detection \cite{Tran:2015} \cite{litjens2017survey}. Roth et al. \cite{Roth:2016} and Prasoon et al. \cite{Prasoon:2013} adapted 2D CNNs for processing 3D volumetric data. However, these studies reported having difficulties when attempting to employ 3D CNN on their data, since they often lack sufficient training samples and computational resources to learn accurate 3D models. Although some authors \cite{deBrebisson:2015} \cite{Urban:2014} have utilized 3D CNNs to process medical images, their architectural settings, convolution kernels, and prediction score volumes have not been disclosed in the detail required to reproduce their results \cite{Dou:2016}. Some exceptions, however, include the work of Kamnitsas et al. \cite{kamnitsas2017efficient}, who devised an effective dense training scheme based on 3D CNNs for brain lesion segmentation and dealing with the computational burden of processing 3D medical scans. Moreover, the 3D U-Net architecture of Cicek et al. \cite{cciccek20163d} takes 3D volumes as input and produces volumetric image segmentation. The architecture and data augmentation of the U-net allow learning models with very good generalization performance from only a few annotated samples. Owing to the success of 3D deep neural networks in medical image segmentation, we are motivated to devise an end-to-end network optimization without requiring manual annotations of the visual image quality. Meanwhile, we seek features maximally affected by partial image artefacts, which are also not very sensitive to variability related to the intrinsic anatomy or image modality at hand.

In this paper, we focus on the analysis of short-axis (SA) cine MRI, although the technique can also be generalized to long-axis images. We aim to identify missing apical slices (MAS) and/or basal slices (MBS) in 3D cardiac MRI volumes. In our previous work, we used a 2D CNN constructed on single-slice images and processed them sequentially \cite{Zhang:2016}. However, this solution ignores contextual information contained across slices providing inferior performance compared to a 3D analysis. We assume that 3D CNNs can easily and effectively deal with within-class variability and between-class similarity, which are important sources of the detection error \cite{Cheng:2016}. We seek to learn a feature representation that achieves reliable classification results even with a small amount of training data or a small number of iterations. In this paper, we address incomplete LV coverage detection using a Fisher-discriminative 3D (FD3D) CNN, which utilizes 3D convolution kernels and exploits the spatial contextual information in volumetric data. The proposed FD3D CNN uses the Fisher discriminant criterion \cite{Yang:2014} on the fully connected layer to render features more discriminative and insensitive to geometric structural variations.

To the best of our knowledge, this is the first study tackling the problem of automatic detection of missing basal and apical slices on a CMR dataset as extensive and challenging as the UK Biobank. Besides introducing a novel FD3D CNN architecture, we propose an effective cascaded detection strategy for incomplete coverage identification. In the first stage, we train two separate FD3D CNN classifiers to detect the absence of basal and apical slices.
In the second stage, we combine the classification results from stage 1 to determine the type of incomplete coverage found on the image.

The rest of this paper is organized as follows. Section \ref{Full} introduces the proposed FD3D CNN architecture and explains the learning strategy for its parameters. Section \ref{Materials} presents experimental materials and metrics. Section \ref{Experiments} describes the experimental design and classification results. Further analysis and discussion of the proposed method are provided in Section \ref{Discussion}. Conclusions are presented in Section \ref{Conclusion}.

\vspace{1em}
\section{Full LV Coverage Detection Method} \label{Full}
\vspace{1em}
\subsection{Problem Formulation}

\begin{figure}	
	\centering	
	\includegraphics[scale=0.37]{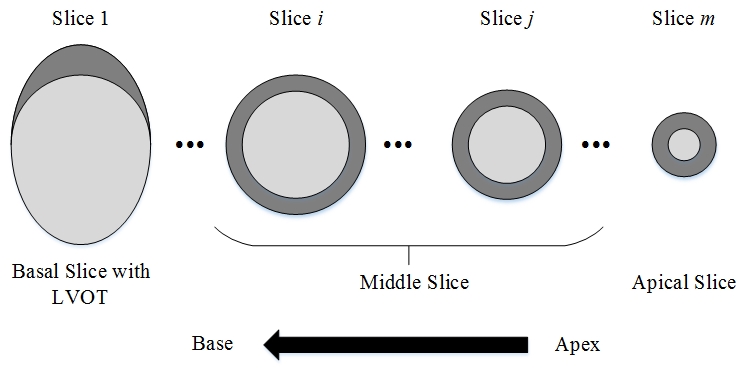} 
	\caption{Schematic LV shapes showing blood pool (light grey) and myocardium (dark grey) for different slices from apex to base. Slice 1 (left) shows LVOT, which identifies the basal slice.}
    \label{figure 2}
\end{figure}

During image acquisition, a sufficient margin ought to be left above and below the LV cavity according to the established guidelines \cite{Schulz-Menger:2013}. However, some image volumes may lack sufficient information at the apical and basal levels, which can hamper or bias the subsequent statistical analysis of cardiac structural and functional parameters in population imaging \cite{Petersen:2017} \cite{Marcus:1999}. In many LV quantification approaches, the LV cross section is approximated using simple quasi-circular models \cite{Childs:2011} \cite{Miller:2013}. These methods can produce a good approximation on LV mid-slices, but not on slices containing the left ventricular outflow tract (LVOT), which is at or near the basal slice. Therefore, in our approach, we treat the blood pool cross-section as a distinct model. Fig. \ref{figure 2} depicts the LV shape of several slices in one cardiac volume from the apex to base.
In volumes with missing basal slice, LVOT is usually not present. 

We use a vector $\mathbf{s}$ to represent pixel values in each slice. A 3D cardiac MRI volume $\mathbf{{V}}$ with full coverage with $n$ slices can be described as follows:

\begin{equation}
	\mathbf{{V}} = \left [ \mathbf{{s}}_1,\mathbf{{s}}_2,...,\mathbf{{s}}_n \right ].
	\label{equation.V}
\end{equation}

Each cardiac volume, $ \mathbf{{V}}=[\mathbf{{s}}_{p},...,{\mathbf{{s}}_{q}}], p\leq q\in [1,n]$, can have a different or same number of slices but cover a different portion of the LV.

To guarantee accurate cardiac volumetry and functional measurements \cite{Klinke:2013}, full LV coverage is a basic requirement \cite{Marcus:1999}. To address this problem, we propose a two stage detection system that first computes image intensity representations by a FD3D CNN model and then detects missing slices based on these representations. In the first stage, we encode spatial contextual information and hierarchically extract high-level features, which indicate intensity representations. Our FD3D CNN model is equipped with a fully connected Fisher discriminative layer (F2) that takes the output of the fully connected layer (F1) as input. In the second stage, independent detection of any missing basal and apical slices is performed and the results are combined to provide the final coverage assessment.

\subsection{Three-dimensional Intensity Representations} 

\begin{figure*}[t]
	\centering	
	\includegraphics[scale=0.44]{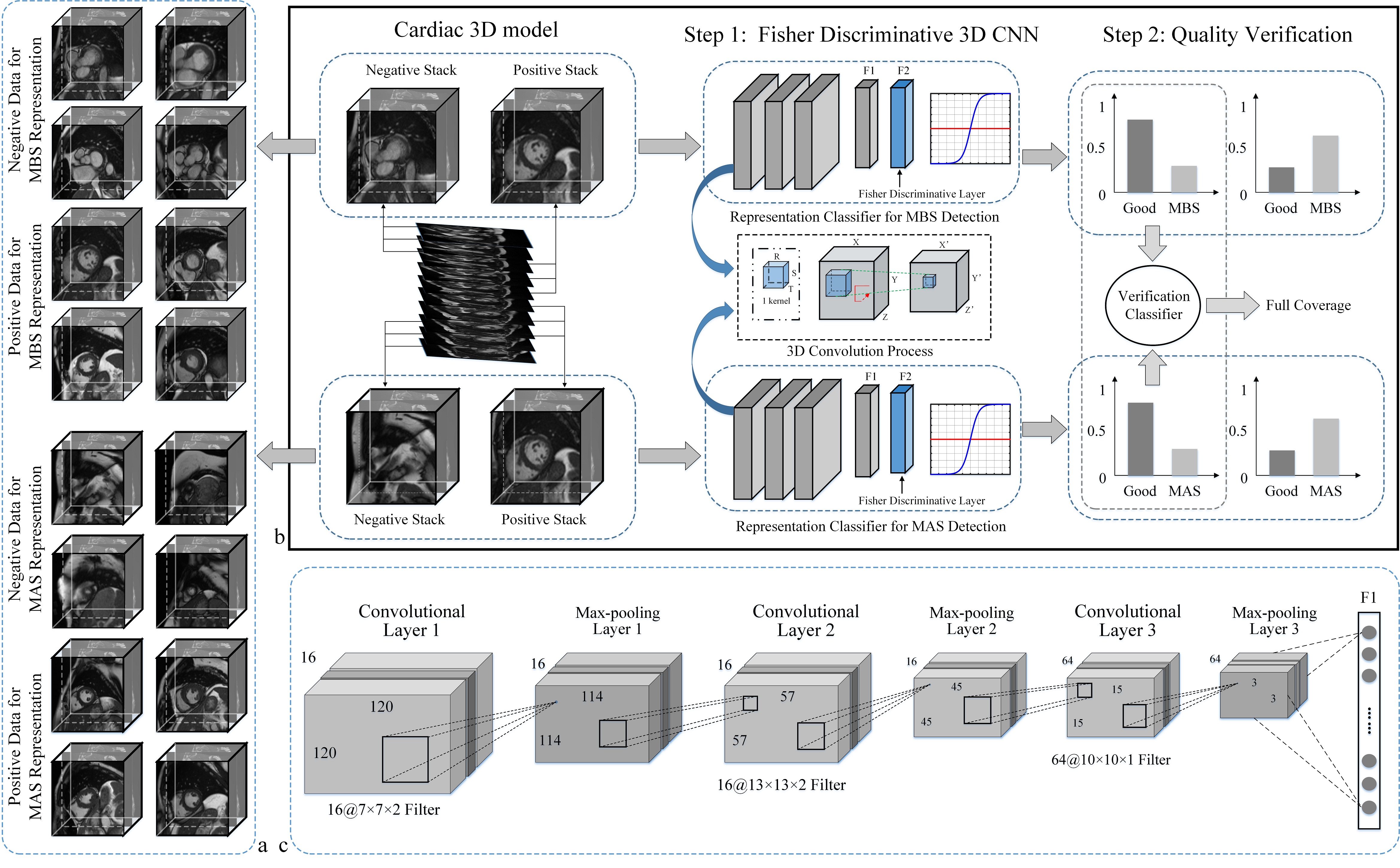} 
	\caption{Whole assessment framework. \textit{a:} Positive and negative training data for each representation classifier (MBS and MAS); \textit{b:} Framework for our LV coverage assessment process; \textit{c:} Structure and parameters of the 3D CNN used in panel b: Step 1.}\vspace{-1em}
	\label{figure 3}
\end{figure*}

Lu et al. \cite{Lu:2009} proposed a pattern recognition technique built on intra-segment correlation, using a normalization scheme, which maps each LV slice to polar coordinates with fixed size, shape level, and position. Intensity information and slice position are relevant even with incomplete LV coverage detection. In our paper, we define intensity representation for the missing slice in a high-level feature space where slices of cardiac MRI are used to construct a representation of intensity. Each slice of the 3D volume is accounted for and the similarity of neighboring slices determines the difference of the 3D intensity distribution. Different characteristics in each slice and contextual information about spatial relation between slices are used to compute intensity representations.

\textit{Which 3D intensity representations?} Our intensity representations are computed as a feature distribution matrix, which integrates information about LV shape and size. We detect incomplete LV coverage by image classification using the distribution matrix. We define two classes: missing apical slice (MAS) and missing basal slice (MBS). 

Given a particular describable visual representation, we can formalize our notion of 3D intensity representations based on Eq. \ref{equation.V}. For example, if we are looking at the volume from base to apex, MAS and MBS can be formalized as follows:

\begin{equation}
\left\{
\begin{array}{l}
\mathbf{{V}}_{MBS}=[\mathbf{{s}}_{q},...,\mathbf{{s}}_{n}],\\
\mathbf{{V}}_{MAS}=[\mathbf{{s}}_{1},...,\mathbf{{s}}_{p}],
\end{array}
\right.
\end{equation}
where, $p, q\in (1,n)$, $\mathbf{{s}}_1$ is the basal slice and $\mathbf{{s}}_n$ is the apical slice. Our intensity representations classifiers can be thought of functions $ f\left( \cdot  \right) $ for mapping 3D stacks $ \mathbf{{V}} $ to real value $p_i $. A positive value of $ p_i $ indicates the presence or strength of the $i^{th}$ representation, while negative values indicate its absence. Considering our intensity representations, if we define $ \mathbf{{V}}_1 $ and $ \mathbf{{V}}_2 $ as MBS and non-MBS samples, respectively, the representation function $ f_{MBS}\left( \cdot  \right) $ may map $ \mathbf{{V}}_1 $ to a positive value and $ \mathbf{{V}}_2 $ to a negative value. This is a binary classification function. Our 3D intensity representation classifiers are trained on the UK Biobank dataset as they provide reliable ground-truth labels based on visual inspection and manual annotation.

\subsection{Fisher Discriminative 3D CNN Model}
In this subsection, we propose a FD3D CNN (shown in Fig. \ref{figure 3}b) to extract high-level features, which represent 3D intensity representations. Our FD3D CNN model is designed by adding a new Fisher-discriminative fully connected layer, F2, which uses the output of the previous layer, F1, as input. The new layer is then stacked onto a conventional 3D CNN. To maximize inter-class distances between learned features while minimizing intra-class distances of learned features, 
we train the newly added Fisher discriminative layer F2 on CNN features based on a  Fisher discriminant criterion \cite{Yang:2014}.

\textit{1) 3D CNN:} Learning feature representations in three dimensions is important for later feature detection and image interpretation tasks in volumetric medical imagery. We employ 3D convolution kernels to encode richer spatial information in volumetric data. Here, feature maps are 3D blocks instead of 2D patches. Conventional 3D convolution is achieved by convolving a 3D kernel, with the cube formed by stacking multiple contiguous slices. With this construction, feature maps in the convolution layer are connected to multiple contiguous frames of the previous layer \cite{Ji:2013} \cite{Huang:2007}. Given an input $ \textbf{v}_k^l$, the 3D convolution layer output equates to a filtering operation with a filter ${\mathbf{{W}}_{ik}^{l+1}}$. Computation of the 3D feature volume $\textbf{h}_i^{l+1}$ is given by: 
\begin{equation}
{\bf{h}}_i^{l + 1} = f\left( {\sum\limits_k {\sum\limits_{r = 0}^{R - 1} {\sum\limits_{s = 0}^{S - 1} {\sum\limits_{t = 0}^{T - 1} {{\bf{W}}_{ik}^{l + 1}(r,s,t){\bf{v}}_k^l + b_k^{l + 1}} } } } } \right)
\label{3Dconv}
\end{equation}
where ${\mathbf{{W}}_{ik}^{l+1}}(r, s, t )$ is the element-wise weight in the 3D convolution kernel, $ {\mathbf{{W}}_{ik}^{l+1}} $ and $b_k^{l+1}$ are the filter and bias terms connecting the feature maps of adjacent layers, and $f( \cdot )$ is the element-wise, non-linear activation function.

\textit{2) Fisher Discriminative 3D CNN:} To boost the discriminative power of 3D CNN learned features, we impose a Fisher discrimination criterion \cite{Yang:2014} on them. Given the 3D input data $ \textbf{V}_{i}^{t}$, where $i$ is the representation class, with $ i = \{1, 2\} $, corresponding to MAS and MBS; the superscript $t$ in $ \textbf{V}_{i}^{t}$ indicates whether the representation is positive or negative, i.e., $ t = \{0, 1\} $; $ \textbf{V}_{i}^{t}=\left [ \textbf{v}_{i,1}^{t}, \textbf{v}_{i,2}^{t},..., \textbf{v}_{i,C}^{t} \right ] $, $\textbf{v}_{i,j}^{t}$ is the input data of $j^{th}$ sample from class $i$, for $j=1,2,...,C$. We denote $ \mathbf{F}_{i,j}^{t} $ to be features in the fully-connected layer of the 3D CNN for class $ i $ and $j^{th}$ sample. Using the Fisher criterion, discrimination is achieved by minimizing within-class scatter of $ \mathbf{F}^{t} $, denoted by $ S_{w}(\mathbf{F}^{t}) $, and maximizing between-class scatter of $ \mathbf{F}^{t} $, denoted by $ S_{b}(\mathbf{F}^{t}) $. $ S_{w}(\mathbf{F}^{t}) $ and $ S_{b}(\mathbf{F}^{t}) $ are defined as follows:
\begin{equation}
S_{w}(\mathbf{F}^t)=\sum_{i=1}^{I}\sum_{\mathbf{F}_{i,j}^{t}\in t}(\mathbf{F}_{i,j}^{t}-\textbf{m}_{i}^{t})(\mathbf{F}_{i,j}^{t}-\textbf{m}_{i}^{t})^{T},
\end{equation}

\begin{equation}
S_{b}(\mathbf{F}^t)=\sum_{i=1}^{I}n_i(\textbf{m}_{i}^{t}-\textbf{m}^{t})(\textbf{m}_{i}^{t}-\textbf{m}^{t})^{T},
\end{equation}
where $\textbf{m}_{i}^{t}$ and $\textbf{m}^{t} $ are mean vectors of $\mathbf{F}_{i}^{t}$ and $\mathbf{F}^{t}$, respectively, and $ n_i $ is the number of samples from class $ i $. The Fisher discriminant regularization term $\Phi(\mathbf{F}^t)$ is defined as $\tr(S_w(\mathbf{F}^t))-\tr(S_b(\mathbf{F}^t))$. To obtain a discriminative classification result with deep learned features, we propose modifying the objective function of the FD3D CNN model by inserting a Fisher discriminant regularization term:
\begin{equation}
\begin{aligned}
{{\mathbf{J}}^*}({\mathbf{W}},{\mathbf{b}}) = \mathop {\arg \min }\limits_{{\mathbf{W}},{\mathbf{b}}} \frac{1}{I}\sum\limits_{i = 1}^I {{y^t}\log a({\mathbf{V}}_{i,j}^t;{\mathbf{W}},{\mathbf{b}})} 
\\
+ (1 - {y^t}) \log (1 - a({\mathbf{V}}_{i,j}^t;{\mathbf{W}},{\mathbf{b}})) 
\\
+ \frac{1}{2}\lambda \left\| {{{\mathbf{W}}}} \right\|_2^2 + \frac{1}{2}\eta {\mkern 1mu} (tr({S_w}({{\mathbf{F}}^t})) - tr({S_b}({{\mathbf{F}}^t}))),
\end{aligned}
\label{equation_d3dcnn}
\end{equation}
where $ \mathbf{{J}}^* $ is our new cost function that can minimize within-class scatter and maximize between-class scatter, and $y$ is the output label. Output activation $ a({\mathbf{V}}_{i,j}^t;{\mathbf{W}},{\mathbf{b}}) = 1/(1 + {e^{ - \textbf{W}{\mathbf{V}}_{i,j}^t - \textbf{b}}}) $ is typically restricted to the open interval $(0,1)$ by using a logistic sigmoid, which is parametrized by $ \mathbf{{W}} $ and $ \mathbf{{b}} $ on the $ j^{th} $ training sample. $\left\| {{\textbf{W}}} \right\|_2^2$ is a penalty term to the loss function that prevents weights from getting too large and helps to prevent over-fitting. Weights in each layer can be adjusted toward target classes and utilize input data close to the corresponding classes in case of no large dataset or a small number of iteration. Here, $\lambda, \eta \in [0,1]$ are two trade-off parameters that control the relative importance of each term and are usually chosen by experiments, which can differ depending on different databases and network structures.

For intensity representation $\mathbf{{V}}_{i,j}^t$, we define:
\begin{equation}
\begin{aligned}
\mathbf{{J}}(\mathbf{{W}},\mathbf{{b}})=&{{y^t}\log a({\mathbf{V}}_{i,j}^t;{\mathbf{W}},{\mathbf{b}})} 
\\
&+ (1 - {y^t}) \log (1 - a({\mathbf{V}}_{i,j}^t;{\mathbf{W}},{\mathbf{b}})),
\end{aligned}
\label{equation 11}
\end{equation}

\begin{equation}
\begin{aligned}
\Phi(\mathbf{F}^t)=\frac{1}{2}\tr((\mathbf{F}_{i,j}^{t}-\textbf{m}_{i}^{t})(\mathbf{F}_{i,j}^{t}-\textbf{m}_{i}^{t})^{T})
\\
-\frac{1}{2}\tr((\textbf{m}_{i}^{t}-\textbf{m}^{t})(\textbf{m}_{i}^{t}-\textbf{m}^{t})^{T}).
\end{aligned}
\label{equation 12}
\end{equation}

Once the new cost function is obtained, we can employ the gradient descent method \cite{Krizhevsky:2012} to solve this optimization problem. Our key problem is to calculate the error of output units, which consists of output errors from two sub-functions $\mathbf{{J}}(\mathbf{{W}},\mathbf{{b}})$ and $\Phi(\mathbf{F}^t)$. To update parameters $ \mathbf{{W}}^t $ and $ \mathbf{{b}}^t $, we first calculate the error $ \delta _i^{L,t} $ (L is the output layer) of the output layer with forward propagation, and then adopt the back-propagation method \cite{Ionescu:2015} to calculate the error $ \delta_i^{l,t} (l < L) $ for other layers. Partial derivatives of the overall cost function $\mathbf{{J}}^*(\mathbf{{W}},\mathbf{{b}})$ regarding $ \mathbf{{W}}^t $ and $ \mathbf{{b}}^t $ are:
\begin{equation}
\begin{aligned}
	\frac{{\partial {\textbf{J}^*}(\textbf{W},\textbf{b})}}{{\partial {W^{l,t}}}} = \sum\limits_{t = 0}^C {\sum\limits_{{F^t} \in t} {\frac{{\partial \textbf{J}({\textbf{W}^t},{\textbf{b}^t})}}{{\partial {W^{l,t}}}}} }  + \eta \sum\limits_{t = 0}^C {\sum\limits_{{F^t} \in t} {\frac{{\partial \Phi ({\textbf{F}^t})}}{{\partial W}}} },
\end{aligned}
\label{equation 17}
\end{equation}

\begin{equation}
\begin{aligned}
   \frac{{\partial {\textbf{J}^*}(\textbf{W},\textbf{b})}}{{\partial {b^{l,t}}}} = \sum\limits_{t = 0}^C {\sum\limits_{{F^t} \in t} {\frac{{\partial \textbf{J}({\textbf{W}^t},{\textbf{b}^t})}}{{\partial {b^{l,t}}}}} }  + \eta \sum\limits_{t = 0}^C {\sum\limits_{{F^t} \in t} {\frac{{\partial \Phi ({\textbf{F}^t})}}{{\partial b}}} }.
\end{aligned}
\label{equation 16}
\end{equation}

In this stage, we use the 3D CNN model with architecture in Table \ref{table 1}. Algorithm 1 provides the pseudo-code to train this new network. In our 3D CNN implementation, a rectifier linear unit (ReLU) \cite{LeCun:2015} is utilized as a non-linear activation function in layers C and F1.
 
\begin{algorithm}
	\caption{FD3D CNN Training.}
	\KwIn{input-target pairs ($ \mathbf{{v}}_{i,j}^t $, $ \mathbf{{y}}^t $), corresponding $j^{th}$ pairs from class $i$, $t$ indicates positive or negative sample; $\eta$.}
	\KwOut{FD3D CNN weight and biases, respectively, $ \mathbf{{W}}=[\mathbf{{W}}^{1,t},\mathbf{{W}}^{2,t},...,\mathbf{{W}}^{l,t}] $ and $ \mathbf{{b}}=[\mathbf{{b}}^{1,t},\mathbf{{b}}^{2,t},...,\mathbf{{b}}^{l,t}] $.}

	\textbf{Begin}\\
	Initialize $ \mathbf{{W}}_{i,j}^t $ and $ \mathbf{{b}}_{i,j}^t $\\
	\While{stopping criterion has not been met}
	{
    \small
		1) Classification error: $ \argmin_{\mathbf{W},\mathbf{b}}\sum_{i=1}^{\rm I} {{y^t}\log a({\mathbf{V}}_{i,j}^t;{\mathbf{W}},{\mathbf{b}})} + (1 - {y^t}) \log (1 - a({\mathbf{V}}_{i,j}^t;{\mathbf{W}},{\mathbf{b}}))  $.\\
		2) Fisher discriminant: $ \Phi(\mathbf{F}^t)=\tr(S_w(\mathbf{F}^t))-\tr(S_b(\mathbf{F}^t)) $.\\
		3) Discriminative objective function: $\argmin_{\mathbf{W},\mathbf{b}}\sum_{i=1}^{\rm I}{{y^t}\log a({\mathbf{V}}_{i,j}^t;{\mathbf{W}},{\mathbf{b}})} + (1 - {y^t}) \log (1 - a({\mathbf{V}}_{i,j}^t;{\mathbf{W}},{\mathbf{b}}))+\frac{1}{2}\lambda \left\| {{\textbf{W}^t}} \right\|_2^2+\frac{1}{2}\eta\,\Phi$.\\
		4) Update $ \mathbf{{W}}_{i,j}^t $ and $ \mathbf{{b}}_{i,j}^t $ with Eqs. (\ref{equation 17}) and (\ref{equation 16}).
	}
	\Return $ \mathbf{{W}}_{i,j}^t $ and $ \mathbf{{b}}_{i,j}^t $ until values of $ \mathbf{{J}}^*(\mathbf{{W}},\mathbf{{b}}) $ in successive iterations are close enough or the maximum number of iterations is reached.\\
	\textbf{End begin}
\end{algorithm}

\begin{table}[t]
	\centering
	\caption{Architecture of the 3D Discriminative CNN Model}
	\begin{tabular}{ccccc}
		\toprule
		Layer & Kernel Size & Stride & Output size & Feature volumes \\
		\midrule
		Input & \textendash    & \textendash     & $ 120 \times 120 \times 3 $ & 1 \\
		C1    & $ 7\times7\times2 $ & 1     & $ 114 \times 114 \times 2 $ & 16 \\
		M1    & $ 2 \times2 \times 1 $ & 2     & $ 57 \times 57 \times 2 $ & 16 \\
		C2    & $ 13 \times 13 \times2 $ & 1     & $ 45 \times 45 \times 1 $ & 16 \\
		M2    & $ 3\times3\times1 $ & 1     & $ 15\times15\times1 $ & 16 \\
		C3    & $ 10\times10\times1 $ & 1     & $ 6\times6\times1 $ & 64 \\
		M3    & $ 2\times2\times1 $ & 1     & $ 3\times3\times1 $ & 64 \\
		F1    & \textendash      & 1     & $ 1\times1\times1 $      & 256 \\
		F2    & \textendash      & 1     & $ 1\times1\times1 $      & 4 \\
		\bottomrule
	\end{tabular}%
    \begin{tablenotes}
     \item Note: F2 is the Fisher Discriminant Layer.
    \end{tablenotes}
	\label{table 1}%
\end{table}%

\section{Materials and Metrics} \label{Materials}
\vspace{0.5em}
\subsection{CMR Acquisition Protocol and Annotation}

\textit{UK Biobank CMR Protocol:} UK Biobank's CMR acquisitions are performed on a clinical wide bore 1.5T scanner (MAGNETOM Aera, Syngo Platform VD13A; Siemens Healthcare, Erlangen, Germany) and include piloting, sagittal, transverse, and coronal partial coverage of the chest and abdomen. For measuring the cardiac function, three long-axis cines are acquired (viz. horizontal long-axis (HLA), vertical long-axis (VLA), and LVOT in both sagittal and coronal views). In addition, a complete SA stack is acquired. All acquisitions use balanced steady-state free precession (bSSFP) MRI sequences, attempting full coverage of the LV and right ventricle \cite{petersen2015uk}. In this study, we will focus on SA bSSFP cine CMR data. To date, more than 18,800 volunteers have been scanned. Voxel and matrix size of these CMR images are, respectively,  $1.8\times 1.8\times 8.0 \textnormal{mm}^3$, and $208 \times 187$ with, approximately, 10 slices per volume. Each volumetric sequence contains about 50 cardiac phases.

\textit{Gold-standard image quality annotations:} Quality-scored cardiac MRI data are available for approximately 5,000 volunteers of the UK Biobank (UKBB) imaging resource. Following visual inspection, manual annotation was carried out with a simple three-grade quality score \cite{carapella2016towards}: (1) optimal quality for diagnosis, (2) suboptimal quality yet analysable and (3) bad quality and diagnostically unusable. In 5,065 SA cine CMR from the same number of volunteers, 4,361 sequences correspond to a quality score of 1, an additional 527 sequences have a quality score of 2, and the remaining 177 sequences have a quality score of 3. All datasets with optimal quality (score 1) had full coverage of the heart from base (LVOT existing) to apex (LV cavity still visible at end-systole). These data were used to construct  the ground-truth classes for our experiments. Note that having full coverage should not be confused with having top/bottom slices corresponding exactly to the base/apex. 

\subsection{Training and Testing Set Definitions}

\textit{Training set:} To create a training dataset for learning intensity representations, we extract the three topmost slices as negative samples for MBS detection (i.e. containing the cardiac base), and the three bottom most slices as negative samples for MAS detection. To create positive samples (i.e. not containing the cardiac base/apex), we choose three-slice blocks, each starting from the middle slice towards the base/apex for MBS/MAS detection training. We create the training set from images with optimal quality and with exclusively full coverage.

We train using three-slice stacks (or triplets) to model the 3D context. the average number of slices per image volume is approximately 10. During training, we extract four triplets (two samples including base/apex and two samples excluding the base/apex). To maximize inter-class separation, it is wise to avoid intersection between the training samples; for example, if we use four-slice stacks (for a ten-slice volume), there will be a two-slice overlap between basal positive/negative examples and the apical region. By choosing the proposed slice triplets, we ensure that there is no overlap and increase the discriminative power of the FD3D CNN.
Another important observation that supports the choice of slice triplets is that the CMR scan volume is not acquired immediately. Instead, each slice is collected over several cardiac cycles leading to some degree of slice-to-slice misalignment. This effect is minimized when considering only slice triplets in contrast to using the full 3D volume.

\textit{Testing set:} During testing, we extract every set of three adjacent slices from top to bottom for each volume and apply these triplets to intensity representation classifiers. Data with known MBS/MAS are created by manually removing the three topmost/bottom most slices from images with optimal quality, as in the training set.

During training and testing, three-slice stacks are input to the proposed FD3D CNN. Scores of the output layer can be interpreted as the probability that triplets correspond to negative or positive MBS/MAS. The final output is the combination of two CNN outputs (MBS and MAS). The three slice stacks input into our network are cropped centered images of dimensions $120 \times 120 \times 3$ to extract the region of interest. Parameter setting of block-size determination is explained in Section~\ref{performance}.

\subsection{Training Set Augmentation}
To prevent over-fitting due to insufficient training data and to improve the detection rate of our algorithm, we employ data augmentation techniques to artificially enlarge our dataset \cite{hoogi2017adaptive} \cite{mortazi2017multi}. 
In our application, we augment the data by applying a discrete set of in-plane rotations and isotropic scalings to the training images. Unlike data augmentation choices made for natural image datasets where variability in location and pose of objects are relatively high, our data are comparatively constrained due to standard imaging protocols and gross patient positioning on the MRI scanner. We therefore chose a set of realistic rotations and scaling factors for MRI. Based on analysis of the in-plane orientation angle distribution for 5,000 subjects for which manual segmentations were available (and therefore LVRV angle can be computed), we found that LVRV orientation ranges between $-45^{\circ}$ and $45^{\circ}$. The set of rotations chosen was accordingly $-45^{\circ}$ and $45^{\circ}$, with two scaling factors of 0.75 and 1.25. This increases the number of training samples by a factor of four, while not adding significantly to the convergence time.

After data augmentation, we constructed 845,000 3D stacks comprised of 2D CMR slices from 3,380 sequences each with 50 cardiac phases, with a quality score of 1. These data are used for experiments in Section IV-A, B, and C. We set aside 981 sequences and data with quality scores of 2 and 3 for later use, as described Section IV-D. In our experiments, 10-fold cross-validation \cite{Kumar:2011} was used to evaluate the performance of our system. To the  best of our knowledge, this is the largest annotated dataset available to date for automatic CMR quality assessment.

\subsection{Learning Performance Metrics}
To evaluate the learning process, we use the following established classification metrics:
\begin{eqnarray}
	\textrm{Precision} = {TP}/{(TP + FP)},\\
    \textrm{Sensitivity} = {TP}/{(TP + FN)},\\
    \textrm{Error Rate} = (FP+FN)/N,
\end{eqnarray}
where $ TP $, $ FP $ and $ FN $ are numbers of true-positive, false-positive and false-negative samples, respectively, and $N$ represents the number of subjects in the test set.

\section{Experiments and Results} \label{Experiments}

\subsection{Performance Analysis} 
\label{performance}
We experiment to characterize the performance of our FD3D CNN learning framework. The error (cost) functions used in learning (Eqs. [\ref{equation_d3dcnn}] and [\ref{equation 11}]) remain within this range $[0, 1]$. In all experiments, the learning process was terminated when standard deviation of the error function over the last five iterations is smaller than $\sigma =0.01$.

\begin{figure}
    \centering	
    \subfigure[]{\includegraphics[scale=0.34]{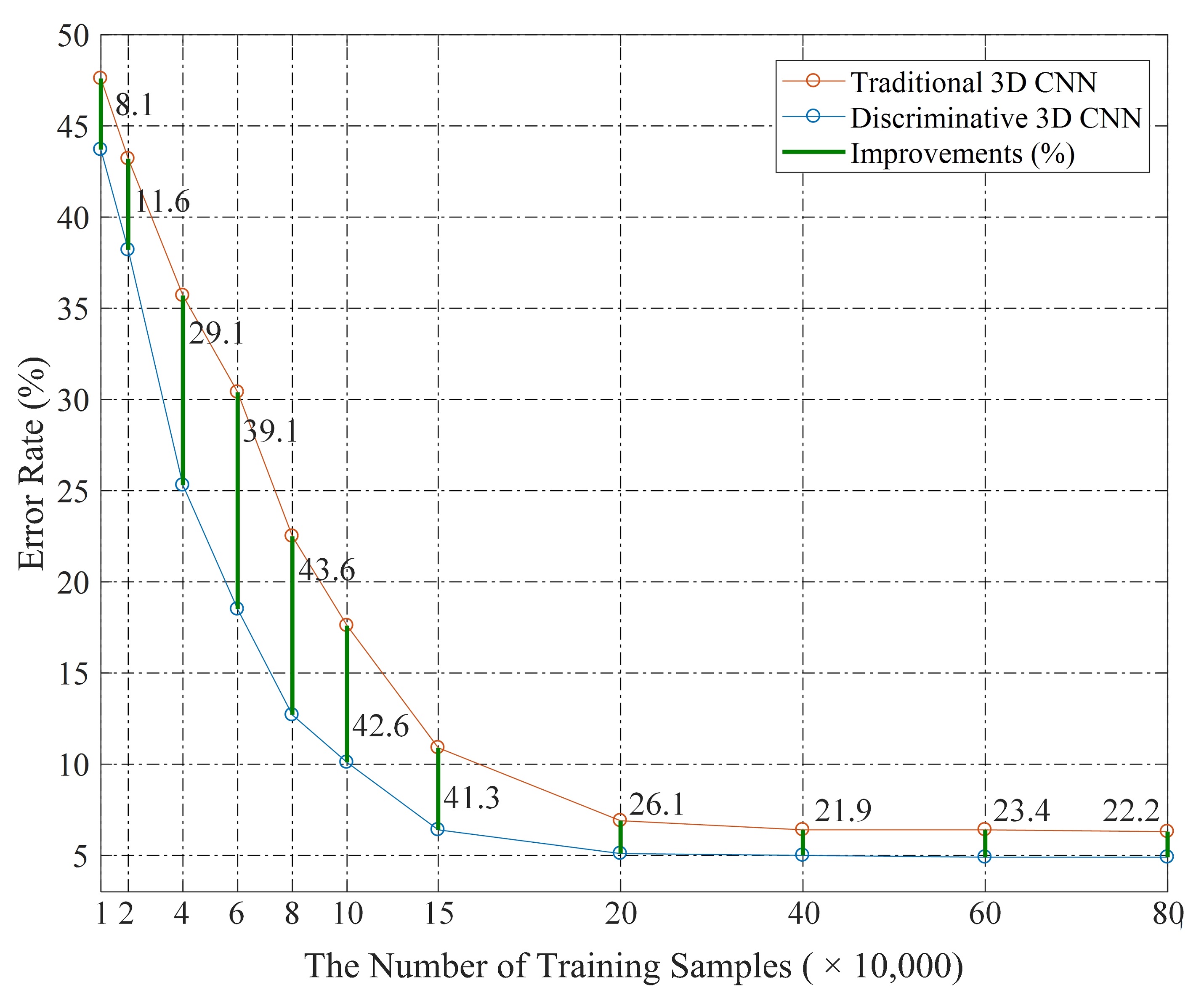}}	
    \subfigure[]{\includegraphics[scale=0.34]{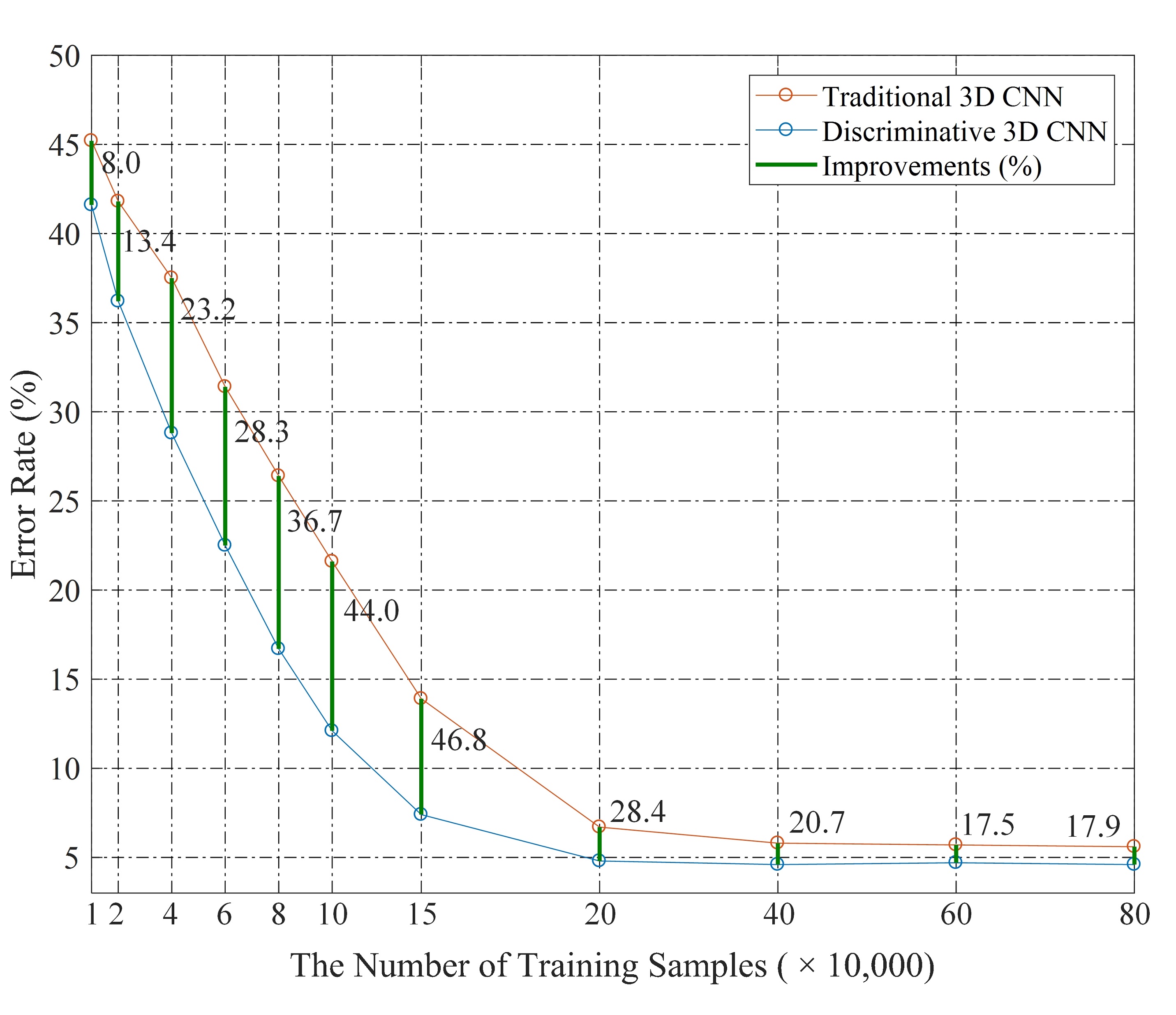}}
	\caption{Error rates and improvements for increasingly larger training sets: (a) MBS detection, (b) MAS detection.}
	\label{figure 4}
\end{figure}

\textit{1) Hyper-parameter selection:} LeCun \textit{et al.} \cite{Sermanet:2012} and Salah \textit{et al.} \cite{Salah:2002} used CNN to recognize handwritten digital numbers with different numbers of training samples on the MNIST dataset. Their results illustrated that, when reducing training samples, the recognition rate of the algorithm drops sharply. To demonstrate the behaviour of our FD3D CNN, we experiment with different percentages of training samples. We use \textit{improvement} defined as $(1-ER_{\rm D}/ER_{\rm T})\times 100$ to benchmark our method against a traditional 3D CNN, where $ER_{\rm D}$ and $ER_{\rm T}$ are error rates of our FD3D CNN and the traditional 3D CNN, respectively. Error rates of MBS/MAS representation learning are shown in Fig. \ref{figure 4}, where our proposed method appears to achieve comparable results with less training data compared to the conventional 3D CNN. We choose $80\%$ of the 845,000 as the training samples and perform testing on the remaining $20\%$. the results are shown in Table \ref{table 6}. Even when trained with fewer iterations, our method achieves better results than the traditional 3D CNN.

\begin{table}[ht]
	\newcommand{\tabincell}[2]{\begin{tabular}{@{}#1@{}}#2\end{tabular}}
	\centering
	\caption{Error rates versus Learning epochs}
	\begin{tabular}{cccc}
		\toprule
		\multirow{3}[4]{*}{\tabincell{c}{Epochs}} & \multicolumn{2}{c}{Error Rate (\%)} & \multirow{3}[4]{*}{\tabincell{c}{Improvement (\%) \\ (MBS/MAS)}} \\	
		& \tabincell{c}{Traditional \\ 3D CNN \\ (MBS/MAS)} & \tabincell{c}{Discriminative \\ 3D CNN \\ (MBS/MAS)} &  \\
		\midrule
		1     &  32.4/30.7   &  28.8/27.4  &  11.1/10.8\\
		10    &  25.4/24.2   &  19.2/17.6  &  24.4/27.3\\
		20    &  19.2/18.7   &   11.3/10.8   &  41.1/42.2\\
		30    &  12.7/13.1     &   8.3/8.6   &  34.6/34.4\\
		40    &  6.3/5.6     &  4.9/4.6    &  22.2/17.9  \\
		\bottomrule
	\end{tabular}%
	\label{table 6}%
\end{table}%

With sufficient training samples and iterations, most machine learning methods can improve their accuracy at a higher computational cost. However, we usually want to obtain a trained network as quickly as possible. This is especially important in population imaging as new datasets can become available and retraining might be required. Rapid training is also a desirable feature during algorithmic development since finding an optimal architecture may require multiple training procedures for different parameter settings. We illustrate that our FD3D CNN has better error-reducting performance as a function of the number of training samples and iterations than other competing techniques.

\begin{figure*}[t]
	\centering
	\subfigure[Sample volumes for MBS testing with automatic quality (AQ), expert cardiologist (VQ1) and cardiac image expert's visual (VQ2) qualities.]{
		\label{fig9:subfig:a} 
		\includegraphics[scale=0.27]{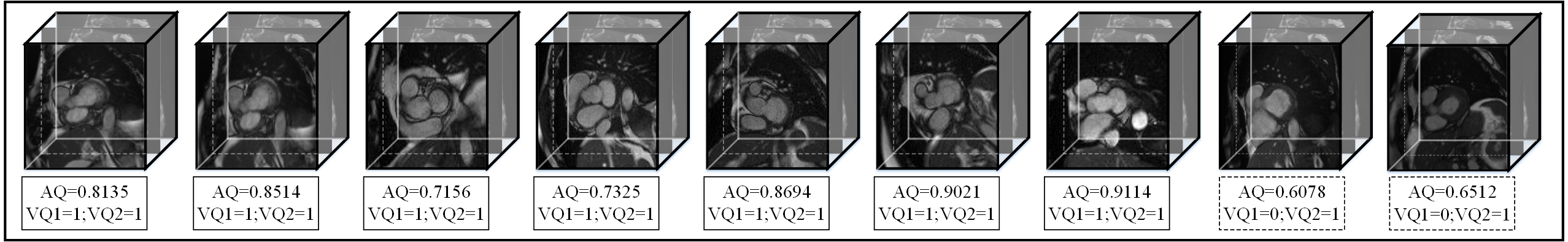}}
	\hspace{1in} \subfigure[Sample volumes for MAS testing with automatic quality (AQ), expert cardiologist (VQ1) and cardiac image expert's visual (VQ2) qualities.]{
		\label{fig9:subfig:b} 
		\includegraphics[scale=0.27]{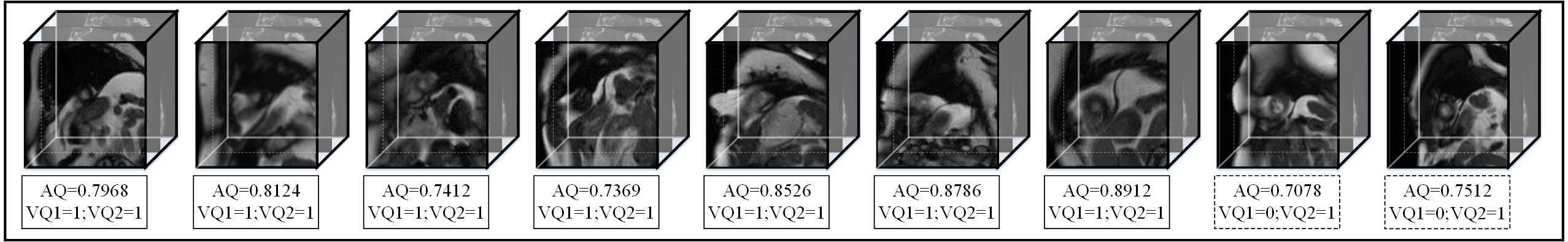}}
	\caption{Sample test volumes and their AQ, expert cardiologist (VQ1) and cardiac image expert's visual (VQ2) qualities for MBS detection (top row) or MAS detection (bottom row) are shown. The left seven samples in each row show consistency between AQ and VQ1, which means our algorithm yields an accurate prediction; The right two samples in each row show the wrong quality prediction and show inconsistency between VQ1 and VQ2.}
	\label{figure 6} 
\end{figure*}

A 3D CNN requires a suitable receptive field (i.e. input size) to achieve the best discrimination. Based on a random sample of 200 image volumes, we determine the smallest crop size that ensures the coverage of the LV structure compared to three block-size configurations, namely, $ 120 \times 120 \times 3 $ (which removes redundant background information based on the central point of original images), $ 180 \times 180 \times 3 $ (which is the original size as extracted and resized from the UK Biobank), and $ 80 \times 80 \times 3 $, which mostly contains the LV at the centre. We test sizes smaller than the original block size of the classification model because we want to determine whether a larger input block with more contextual information can enhance the model's discriminative capacity. The results obtained with these settings are shown in Table \ref{table 4}. With a block size of $80 \times 80 \times 3$, MBS/MAS detection precision rate reaches 89.01\% and 88.36\%, respectively. The detection performance improves to a precision rate of 91.81\% and 90.73\% under block size $120 \times 120 \times 3$, demonstrating that increasing contextual information can enhance the discriminative capacity of 3D CNN. Without cropping, the detection precision rate decreases to 90.12\% and 89.78\% for MBS and MAS detection, respectively. This may have been because too much redundant contextual information clutters the actual LV signature, and hence degrades detection performance. Based on these experiments, we set block size to $120 \times 120 \times 3$, to achieve optimal detection performance. 

\begin{table}[ht]
	\centering
	\caption{Performance versus Block Size}
	\begin{tabular}{ccccc}
		\toprule
		\multicolumn{1}{c}{\multirow{2}[2]{*}{Block Size}} & \multicolumn{2}{c}{Precision} & \multicolumn{2}{c}{Sensitivity} \\		
		\multicolumn{1}{c}{} & MAS   & MBS   & MAS   & MBS \\
		\midrule
		$ 80 \times 80 \times 3 $ & 89.01\% & 88.36\% & 88.24\% & 87.94\% \\
		$ 120 \times 120 \times 3 $ & \textbf{91.81\%} & \textbf{90.73\%} & \textbf{90.92\%} & \textbf{90.25\%} \\
		$ 180 \times 180 \times 3 $ & 90.12\% & 89.78\% & 89.63\% & 88.92\% \\
		\bottomrule
	\end{tabular}%
	\label{table 4}%
\end{table}%


Typical classification results using the proposed FD3D CNN architecture are shown in Fig. \ref{figure 6}. A few basal stacks (top row) and apical stacks (bottom row) in the test datasets with their AQ or corresponding posterior probability values are shown. High score values on the stack correspond to the likelihood of being a correct basal or apical triplet. Basal slices with existing LVOT indicate higher probability values of being correctly classified. This shows that the training captures the LVOT as a prominent feature in correctly positioned basal slices.

\textit{2) Comparison to other machine learning methods:} We compare our framework with a traditional 3D CNN and with our previous 2D CNN study \cite{Zhang:2016}. Table \ref{table 2} lists the results for these architectures.The architecture of a traditional 3D CNN is similar to that of our FD3D CNN, replacing the fisher layer (F2) with a traditional fully connected layer including 256 ReLU activation neurons. We use the same training and testing approaches for the 3D CNN and list the results obtained using the hand crafted features used in \cite{Lu:2011}. In \cite{Lu:2011}, the basal slice was identified following these steps: 1. Choose the mid-slice image as the start image and process each image sequentially in the basal direction. 2. Apply the optimal threshold method to convert the ROI to a binary image. 3. Identify the binary object with blood pool, which shows an elliptical shape. 4. Calculate the length of the major axis $L$ of the ellipse that has the same normalized second central moments as the binary object. 5. If the ratio of the current to preceding $L$ exceeds a predefined threshold  (e.g. $>1.2$ in this work), then a basal slice is identified; otherwise, the basal slice is missing. We use a similar method to identify the apical slice. We process each image sequentially from base to apex. If the ratio of the current to preceding $L$ is smaller than a predefined threshold (e.g. $<0.2$ in this study), an apical slice is detected; otherwise, the apical slice is missing. We employ this feature extraction procedure for prediction. The proposed FD3D CNN shows the best precision and sensitivity figures in each representation classifier, and full LV coverage detection performance.

\begin{table*}[ht]
	\centering
    \newcommand{\tabincell}[2]{\begin{tabular}{@{}#1@{}}#2\end{tabular}}
	\caption{Performance comparison of different learning models with learned and hand-crafted visual representations.}
	\begin{tabular}{cccccccc}
		\toprule
		\multicolumn{1}{c}{\multirow{1}[4]{*}{Method}} & \multirow{1}[4]{*}{Features} & \multicolumn{3}{c}{Precision (\%)} & \multicolumn{3}{c}{Sensitivity (\%)}  \\
		\cline{3-8} \\ [0.0pt]
        \multicolumn{1}{c}{} &       & \multicolumn{1}{c}{MAS} & \multicolumn{1}{c}{MBS} & \multicolumn{1}{c}{$\overline{\textrm{MBS} \vee \textrm{MAS}}$} & \multicolumn{1}{c}{MAS} & \multicolumn{1}{c}{MBS} & \multicolumn{1}{c}{$\overline{\textrm{MBS} \vee \textrm{MAS}}$}  \\
		\midrule
		\multicolumn{1}{c}{FD3D CNN} & \multirow{2}[6]{*}{Learned} & {\tabincell{c}{\textbf{91.81} $\pm \textbf{0.21}$ }}& {\tabincell{c}{\textbf{90.73} $\pm \textbf{0.28}$}} &  {\tabincell{c}{\textbf{91.12} $\pm \textbf{0.24}$}} &  {\tabincell{c}{\textbf{90.92} $\pm \textbf{0.26}$}}  & {\tabincell{c}{\textbf{90.25} $\pm \textbf{0.28}$}}   & {\tabincell{c}{\textbf{90.15} $\pm \textbf{0.22}$}} \\
        
		\multicolumn{1}{c}{3D CNN} &   & {\tabincell{c}{89.12 $\pm 0.36$}}  &  {\tabincell{c}{89.32 $\pm 0.34$}}& {\tabincell{c}{89.20 $\pm 0.31$}}  & {\tabincell{c}{89.42 $\pm 0.31$}}  & {\tabincell{c}{89.47 $\pm 0.30$}}  & {\tabincell{c}{89.25 $\pm 0.29$}} \\
        
		\multicolumn{1}{c}{2D CNN} &   & {\tabincell{c}{81.61 $\pm 0.56$}}  &  {\tabincell{c}{74.10 $\pm 0.58$}} & {\tabincell{c}{79.42 $\pm 0.62$}}  & {\tabincell{c}{88.73 $\pm 0.49$}}  & {\tabincell{c}{88.75 $\pm 0.51$}}   & {\tabincell{c}{88.01 $\pm 0.56$}} \\
        
		\midrule
		
		\multicolumn{1}{c}{\tabincell{c}{Lu \textit{et al}. \cite{Lu:2011}}} & Hand-crafted &  {\tabincell{c}{37.60 $\pm 1.22$}}   &   {\tabincell{c}{45.68 $\pm 1.36$}} & {\tabincell{c}{56.92 $\pm 1.71$}}  &   {\tabincell{c}{67.43 $\pm 0.92$}}   & {\tabincell{c}{74.56 $\pm 1.32$}}   & {\tabincell{c}{63.25 $\pm 1.79$}}\\
        
		\bottomrule
	\end{tabular}%
	\label{table 2}%
\end{table*}%



\subsection{Inter-Observer Reliability}
To contextualize the results of automatic full LV coverage assessment, we compare it to the inter-observer full LV coverage detection rate obtained by expert readers. The inter-observer agreement \cite{Gwet:2008} of human experts is evaluated by reassessing a subset of 200 random CMR datasets. The quality distribution levels in this randomly selected subset are compared to original data using Pearson's ${\chi ^2}$ goodness-of-fit test to confirm that it represents the original data distribution ($p>$ 0.05). The reassessed samples demonstrate strong agreement with original qualities (Cohen's $\kappa$ = 0.76, $p$$<$ 0.05).

To show how our results can be compared to the expected human detection error rates, we present the error rates between an expert cardiologist (VQ1) and another cardiac image expert (VQ2) for 200 re assessed samples. The confusion matrix of VQ1 versus VQ2 is presented in Table \ref{tabel VI}. Use of the confusion matrix reveals 7 among the 200 re assessed samples with inconsistent quality assessment between VQ1 and VQ2. These findings show that the expert cardiologist's visual results conflict with the cardiac image expert's visual assessment only 3\% of the time. As shown in Table \ref{table 6} \emph{epoch = 40}, our automatic algorithm's error rate is just below 5\%, which shows excellent agreement with human expert assessments (two percentage points). Some examples of MBS/MAS test images are shown in Fig. \ref{figure 6} (panels a and b correspondingly). We have intentionally chosen to show seven inter-observer agreement examples, plus two disagreement examples on each panel.

\begin{table}
\renewcommand{\arraystretch}{1.4}
\centering
\caption{Confusion matrix of the expert cardiologist (VQ1) and cardiac image expert's visual (VQ2) results. Grey numbers indicate number and ratio of correct estimates.}
\label{my-label}
\begin{tabular}{ll|ccc|c|}
\cline{3-6} 
 &  & \multicolumn{3}{c|}{VQ2}  & \\  
\cline{3-5} &   & MBS  & MAS  & $\overline{\textrm{MBS} \vee \textrm{MAS}}$  &  \multirow{-2}{*}{Correct}    \\ 
\cline{1-6}  
\multicolumn{1}{|c|}{} & MBS  & \cellcolor[rgb]{0.7, 0.7, 0.7}{67} & 0 & 3 &  \cellcolor[rgb]{0.7, 0.7, 0.7}{0.96}   \\   
\multicolumn{1}{|l|}{} & MAS  & 0 & \cellcolor[rgb]{0.7, 0.7, 0.7}{65} & 2 & \cellcolor[rgb]{0.7, 0.7, 0.7}{0.97}    \\  
\multicolumn{1}{|l|}{\multirow{-3}{*}{\begin{sideways}VQ1\end{sideways}}} & $\overline{\textrm{MBS} \vee \textrm{MAS}}$ & 1 & 1 & \cellcolor[rgb]{0.7, 0.7, 0.7}{61} &  \cellcolor[rgb]{0.7, 0.7, 0.7}{0.97}   \\ 
\cline{1-6}
\end{tabular}
\label{tabel VI}
\end{table}

\subsection{Cross-database Performance: Sunnybrook Cardiac Dataset}
We evaluate the generalization of the performance of our full LV coverage detection system on an independent database. We assess the sensitivity of our system to moderate changes in imaging conditions, scanner vendors, image resolution, etc. To this effect, we use Data Science Bowl Cardiac Challenge Data (Kaggle or Sunnybrook Cardiac dataset) \cite{Kaggle:2016}. This dataset comprises 1,120 cardiac MRI volumes. Cine steady state free precession (SSFP) MR short-axis (SAX) images are obtained with a 1.5T GE Signa CV/i MRI System (General Electric, Milwaukee, WI). All images are obtained during 10-15 second breath-holds with a temporal resolution of 20 cardiac phases over the heart cycle (scanned from the ED phase). Six to twelve SAX images are obtained from the atrioventricular ring to the apex (resolution 1.25$\times$1.25$\times$8mm$^3$, thickness = 8mm). Gold-standard full LV coverage is obtained by an experienced reader and checked visually by inspecting slices from base to apex. Original volumes are used for full LV coverage detection and triplets of top and bottom slices are used, respectively, as negative examples for MBS and MAS. Positive examples of MBS/MAS are obtained from triples of mid-slices. This dataset is used as a test set for the FD3D CNN that was pre-trained with 800,000 volumes from the UK Biobank. Values for error, precision and sensitivity under various conditions are shown in Table \ref{table 3}. 

\begin{table}[ht]
	\centering
	\caption{Cross-dataset performance: Kaggle dataset.}
	\begin{tabular}{cccc}
		\toprule
		& Error (\%) & Precision (\%) & Sensitivity (\%) \\
		\midrule
		MAS   &  6.43    &  86.51    & 88.74 \\
        MBS   &  7.02    &  84.03    & 85.69 \\
        $\overline{\textrm{MBS} \vee \textrm{MAS}}$ &  6.64    &  85.74    & 87.01 \\
		\bottomrule
	\end{tabular}%
	\label{table 3}%
\end{table}%

\subsection{Missing Slice Rate per Visual Quality Score}
To gain insight into the relation between missing slice rates and visual quality scores achieved by experts \cite{carapella2016towards}, a third experiment is conducted. The system is trained on 3,380 random volumes from a total of 5,065. The testing set, as earlier indicated, has 1,685 CMR volumes distributed among the quality scores (from 1 to 3: 981, 527 and 177). Table \ref{table 7} gives the percentages of the full LV coverage class for each quality score. CMR data with a quality score of 3 highly correlates with MBS, as missing basal slices highly affect accurate quantitative analysis in CMR.

\begin{table}[ht]
	\centering
    \newcommand{\tabincell}[2]{\begin{tabular}{@{}#1@{}}#2\end{tabular}}
	\caption{Missing Slice Rate per Visual Quality Score.}
	\begin{tabular}{cccccccc}
		\toprule
		Quality Score &  MAS (\%) & MBS (\%) & $\overline{\textrm{MBS} \vee \textrm{MAS}}$ (\%) \\		
		\midrule
		1   &  1.7    &  0.6    & 97.7 \\
		2   &  74.7    &  24.0    & 1.3 \\
        3   &  18.0   &  80.4   &  1.6 \\
		\bottomrule
	\end{tabular}%
	\label{table 7}%
\end{table}%

\vspace{-1em}

\subsection{Clinical Impact}
To assess the impact of incomplete LV coverage in real applications, such as measurement of cardiac function based on blood volumes, we design an experiment where incomplete coverage is simulated and volume differences between full and incomplete coverage are measured. We also compute two commonly used indexes of the cardiac function derived from such volumes viz. stroke volume (SV) and ejection fraction (EF), and similarly report the differences between the full and incomplete coverages. For this experiment, we take 4,737 subjects for which manual annotations are available (both cardiac phase labels and full coverage labels), and systematically remove the basal and apical slices to generate incomplete MBS and MAS volumes. Then, we compute blood pool volumes at the ED and ES phases, and from these, we obtain SV and EF. Finally, the average volumes and indexes are computed across the sample, comparing full coverage and MBS/MAS. Table \ref{Table VIII} shows that the largest effect of incomplete coverage is caused by MBS, where the missing slice reduces ED and ES volumes by an average of 12\% and 20\%, respectively. In turn, these differences cause a decrease in the computed SV by 6.7\% and an increase in the EF by 3.9\%. The absence of the apical slice has a smaller yet non-negligible impact on the volumes and derived indexes.

\begin{table}[ht]
\setlength{\tabcolsep}{1mm}
\renewcommand{\arraystretch}{1.3}
\centering
\newcommand{\tabincell}[2]{\begin{tabular}{@{}#1@{}}#2\end{tabular}}
\caption{Effect of incomplete cardiac coverage (MBS/MAS) on the End-diastolic, End-systolic, stroke volumes and ejection fraction. Values are shown as Mean $\pm$ standard deviations.}
\label{Table VIII}
\begin{tabular}{lr|rr|rr}
\toprule
            & Full  & MBS   & Effect(\%)    & MAS    & Effect(\%)  \\ \midrule
LVEDV(ml)   & 155.8$\pm$35.6 & 136.1$\pm$33.4 & -12.6\%   & 151.5$\pm$35.1  & -2.7\%  \\  
LVESV(ml)   & 66.8$\pm$21.2  & 53.0$\pm$19.0  & -20.0\%   & 64.3$\pm$20.9   & -3.7\%  \\  
LVSV(ml)    & 89.1$\pm$19.8  & 83.1$\pm$19.7  & -6.7\%    & 87.1$\pm$19.6   & -2.2\%  \\
LVEF(\%)    & 57.1$\pm$0.06  & 61.0$\pm$0.06  & +3.9\%    & 57.5$\pm$0.06   & +0.4\%  \\  \bottomrule
\end{tabular}
\end{table}


\subsection{Implementation Considerations}
The experiments reported here are conducted using the ConvNet library \cite{Demyanov:2016} on an Intel Xeon E5-1620 v3 @3.50GHz machine running Windows 10 with 32GB RAM and Nvidia Quadro K620 GPU. The networks are optimized using the gradient descent method \cite{Krizhevsky:2012} with the fllowing hyper-parameters: learning rate = 0.01, momentum = 0.9, drop-out rate = 0.1. Trainable weights are randomly initialized from a Gaussian distribution ($\mu=0$, $\sigma=0.01$) and updated with standard back-propagation. Models converge in about 6 hours when training is performed with 800,000 volumes with size 120$\times$120$\times$3. Testing is rapid and can process each volume in 3 seconds.


\section{Discussion} \label{Discussion}

Automatic identification of CMR volumes with incomplete LV coverage is important in high-throughput image analysis of population imaging. The acquisition of thousands of suboptimal CMR images for later image analysis can be avoided if such quality assessment is performed online and a system provides immediate feedback to technical staff when new images are acquired. Incomplete LV coverage influences the accuracy of anatomical and functional LV parameters of clinical interest. Manual annotation of LV coverage is laborious, time-consuming and error prone in current clinical routines. To automate this labour-intensive task, we propose an efficient and robust two-stage framework for the automatic detection of missing slices at the LV base and apex. In the first stage, we train a FD3D CNN that computes the corresponding intensity representation with high accuracy. It can qualify CMR volumes based on two representations, and can assist radiologists by automatically labelling the potentially incomplete volumes to mark them for closer inspection. The second stage robustly discriminates two quality categories (MBS and MAS), based only on the intensity representation classifiers, which are then used to recognize new cardiac volumes with no further training. Specifically, to use the spatial information in volumetric data, we use 3D CNN with shared 3D convolution kernels. Meanwhile, a Fisher discriminant layer leads to small within-class scatter and large between-class scatter of feature vectors in that layer. Extensive experimental results illustrate the effectiveness and efficiency of our method: its performance is superior to that of other methods with obvious advantages.

In any AIQA system for population imaging, accuracy and robustness are key design criteria. These methods must work without many false positives or false negatives, and must cope with considerable variation in image quality. Most machine learning methods can improve their recognition accuracy by increasing the number of iterations. However, an increasing number of iterations comes at a high computational cost. This can be prohibitive with large databases or when retraining is required as new data become available. In this study, we used a very large dataset comprising more than 5,000 individually annotated cardiac MRI scans of the same number of subjects, which is 50-fold the 100 cases used in our previous study \cite{Zhang:2016}. However, when compared to natural image datasets \cite{Krizhevsky:2012}, our cardiac MRI dataset is still relatively small. We had to design an efficient network taking full advantage of the available data. Considering there were only a few labelled images, there was no point in constructing a network with too many sub-sampling layers; there would have been a higher computational cost with more layers of feature abstraction. Three-dimensional CNN have been among the most promising solutions for object detection tasks. Thus far, most studies have focused on image segmentation and registration, and little effort has been devoted to AIQA. We propose a FD3D CNN with an extra layer using a Fisher discriminant criterion, which tackles the problem of detecting full LV coverage as an important quality criterion. Our method can eliminate redundant convolutional computations during forward propagation and achieve a comparable result with a smaller number of training samples and iterations. Specifically, our FD3D CNN can achieve a high precision rate of nearly 92\%/91\% for MBS/MAS detection with only 20 epochs, which is better than traditional 3D CNN. Meanwhile, even with a small number of training samples (4 $\times$ 10,000), our FD3D CNN can decrease the error rate by approximately 29.1\% compared to traditional 3D CNN approaches for MBS detection.

Our proposed automatic assessment framework for full LV coverage has great potential to improve the robustness of subsequent population image parsing. One can imagine an approach whereby image analysis is adaptive to image quality and where different models are used depending on whether the volume under analysis is missing basal or apical slices. In our architecture, we focus on learning intensity representations and develop a FD3D CNN to describe those that best discriminate the missing apical or basal slices. We then use the computed representation classifiers to identify the final image quality. The advantages of a representation-based method for vision tasks are manifold: they can be composed to create descriptions at various levels of specificity; they are generalizable, as they can be learned once and then applied to recognize new objects or categories with no further training and are efficient, possibly requiring exponentially fewer representations than explicitly naming each category. In the future, we plan to investigate the possibility of detecting full LV coverage for all slices, rather than just for basal/apical slices, so we can directly predict visual quality scores. The difficulty of detecting missing middle slices lies in the similar shape of contiguous LV slices, which makes training the representation classifier a non-trivial task. Another future work is to extend deep-learning methods for multi-plane estimation, that is, regressing one 3D volume to estimating missing slices acquired from different positions. This is a limitation of our two-stage framework, which can only estimate the basal and apical planes. One way to achieve 3D CNN for multi-plane estimation would be to apply regression on each plane separately and then combine all regression results into a single estimation.

\section{Conclusion} \label{Conclusion}

In this study, we tackled the problem of detecting incomplete LV coverage in large population image databases. We illustrated the concept by proposing a Fisher discriminative 3D CNN tested on CMR data from the UK Biobank. Our FD3D CNN was proposed by adding a new Fisher-discriminative fully connected layer into the network, which achieved a significant improvement in intensity representation. The learned representation classifiers were computed for candidates of corresponding quality categories. We also validated our model by training with the UK Biobank dataset and cross-evaluating with data from the Data Science Bowl Cardiac Challenge dataset. The proposed model shows high consistency with human perception and is superior to state-of-the-art methods, showing its high potential. Our proposed FD3D CNN can also be easily applied and boosts results for other detection and segmentation tasks in medical image analysis. 

\section*{Acknowledgements and competing interests}
This research has been conducted using the UK Biobank Resource under Application 2964. The authors wish to thank all UK Biobank participants and staff.

LZ acknowledges funding for his PhD studies from the China Scholarship Council (CSC). MC and AFF acknowledge support from the VPH-DARE@IT FP7 EC Integrated Project (FP7-ICT-2011-9-601055). The EPSRC-funded MedIAN Partnership (EP/N026993/1) also supports AFF. The British Heart Foundation (PG/14/89/31194 also provided partial funding. SKP and SN acknowledge the National Institute for Health Research (NIHR) Oxford Biomedical Research Center based at The Oxford University Hospitals Trust at the University of Oxford, and the British Heart Foundation Center of Research Excellence. SEP acknowledge partial support from the NIHR Biomedical Research Center at Barts and from the SmartHeart EPSRC programme grant (EP/P001009/1). SEP provides consultancy to Circle Cardiovascular Imaging Inc., Calgary, Canada, and the access application 11350 of UK Biobank.

\bibliographystyle{IEEEtran}
\bibliography{ieee-tbme-full-lv-coverage}

\end{document}